\documentclass[conference]{IEEEtran}
\usepackage{cite}
\usepackage{amsmath,amssymb}
\usepackage{graphicx}
\usepackage{booktabs}
\usepackage{multirow}
\usepackage[hyphens]{url}
\title{FAR-DPO: Feasibility-Aware and Robust Direct Preference Optimization for Cyclic Peptide Design}
\author{\IEEEauthorblockN{Guofeng Zhang\textsuperscript{1,2,*}, Rong Han\textsuperscript{3,*}, Xiaoyu Wang\textsuperscript{2}, Zhiyun Li\textsuperscript{1,2}, Zongbo Han\textsuperscript{1}, Xiaohong Liu\textsuperscript{2}, Guangyu Wang\textsuperscript{1,\dag}}
\IEEEauthorblockA{\textsuperscript{1}State Key Laboratory of Networking and Switching Technology,\\
Beijing University of Posts and Telecommunications, Beijing, China\\
\textsuperscript{2}Innovation \& Research Department, South China Hospital, Medical School,\\
Shenzhen University, Shenzhen, Guangdong, China\\
\textsuperscript{3}Department of Computer Science and Technology, Tsinghua University, Beijing, China\\
\textsuperscript{*}Equal contribution\quad \textsuperscript{\dag}Corresponding author: Guangyu Wang (guangyu.wang24@gmail.com)}}

\IEEEoverridecommandlockouts
\IEEEpubid{\begin{minipage}[t]{\textwidth}\footnotesize\raggedright
\rule{\textwidth}{0.4pt}\\[2pt]
Accepted version. \copyright\ 2026 IEEE. Personal use of this material is permitted. Permission from IEEE must be obtained for all other uses, in any current or future media, including reprinting/republishing this material for advertising or promotional purposes, creating new collective works, for resale or redistribution to servers or lists, or reuse of any copyrighted component of this work in other works.
\end{minipage}}

\begin{document}
\bstctlcite{IEEEexample:BSTcontrol}

\maketitle

\begin{abstract}
Cyclic peptides are emerging as promising molecular scaffolds in drug discovery due to their high binding affinity and structural stability. However, extending generative models from linear to cyclic peptide design remains challenging, as cyclization sharply restricts the feasible design space through coupled geometric and biophysical constraints. Moreover, limited training data has led existing approaches to rely largely on zero-shot generation or post hoc filtering, resulting in low yields of feasible designs and limited control over multi-objective trade-offs. To address these limitations, we propose \mbox{FAR-DPO} (Feasibility-Aware and Robust Direct Preference Optimization), an architecture-agnostic framework that steers generative models toward structurally and biophysically feasible cyclic peptide designs, particularly for challenging targets. \mbox{FAR-DPO} integrates feasibility-aware preference construction with difficulty-aware group-robust optimization. Specifically, it constructs within-target preference pairs through feasibility-gated multi-objective dominance and adaptively reweights predefined difficulty groups according to their current preference losses. 
Under a fixed generation budget, \mbox{FAR-DPO} increases overall success rate on the CPSea benchmark from 46.9\% to 57.8\% on PepGLAD and from 48.0\% to 49.6\% on PepFlow. These gains also extend to the hardest target quartile and are accompanied by more favorable best-per-target binding scores. Together, these results demonstrate \mbox{FAR-DPO}’s effectiveness in improving feasibility and target-wise robustness.
\end{abstract}


\section{Introduction}

Cyclic peptides are attracting notable interest as potential therapeutics, particularly due to their unique ability to modulate protein-protein interactions that are often difficult to target with conventional drugs \cite{dougherty2019penetration,villar2014macrocycles}. By constraining the peptide backbone into a closed-ring architecture via terminal or side-chain covalent linkages, cyclization significantly restricts conformational flexibility. This rigidified structure stabilizes bioactive conformations, improves target-binding affinity, and significantly improves resistance to proteolytic degradation \cite{zorzi2017cyclic}.

Recent diffusion- and flow-based generative models can directly co-design peptide sequences and three-dimensional structures conditioned on a target structure \cite{kong2024pepglad,li2024pepflow}, opening new avenues for target-specific cyclic peptide design. However, the generative distributions learned by these models are not explicitly aligned with the coupled structural and biophysical requirements of cyclic peptides. A practically viable design must combine favorable binding with valid ring-closure geometry, correct stereochemistry, favorable interaction energetics, and reasonable steric geometry \cite{bhardwaj2016hyperstable,hosseinzadeh2017macrocycles,yang2025cpsea}. A candidate that performs well on an individual binding metric may therefore still be invalid or unfavorable along other essential dimensions. The central challenge is not merely to recover a few high-scoring candidates through extensive sampling, but to shift the generative distribution toward designs that are jointly feasible and favorable across multiple quality dimensions under a fixed generation budget. Existing approaches improve cyclic peptide design through architecture-specific modeling and multi-stage computational filtering \cite{rettie2025afcyc,rettie2025rfpeptides,jiang2025cpcomposer,tang2024peptune,wang2025cycbuilder}, while recent training-time alignment methods primarily target specific geometric constraints such as ring closure \cite{zhang2026geocycler}. Despite these advances, a general approach for aligning target-conditioned generators with multiple structural and biophysical objectives while maintaining robust performance across heterogeneous targets remains underexplored.

To mitigate this bottleneck, recent efforts have largely followed two distinct paradigms. One line of research incorporates cyclization requirements into generation through specialized architectures or geometric conditioning \cite{rettie2025afcyc,rettie2025rfpeptides,jiang2025cpcomposer}. Another line of research uses inference-time guidance, property optimization, or post hoc filtering to improve candidate quality \cite{tang2024peptune,wang2025cycbuilder}. These approaches have advanced cyclic peptide generation. However, they are often tied to a particular generative architecture or cyclization mechanism. Their optimization primarily focuses on individual properties or specific structural constraints. When cyclization feasibility, structural quality, and target binding must all be satisfied, an improvement in any single metric does not necessarily translate into a higher yield of jointly feasible candidates \cite{jin2020multiobjective,ren2025abnovo}. Moreover, the design difficulty varies substantially across targets, so better aggregate performance does not guarantee consistent gains across targets \cite{duchi2021uniform,sagawa2020groupdro}. Training-time alignment methods, including reward-based fine-tuning and direct preference optimization, offer a complementary way to reshape the generative distribution using scalar rewards or pairwise preferences \cite{olivecrona2017molecular,rafailov2023dpo,wallace2024diffusiondpo}. Nevertheless, a systematic solution is still lacking for aligning a model with multiple feasibility requirements while maintaining robust performance across heterogeneous targets, without relying on a particular generative architecture.

To this end, we propose \mbox{FAR-DPO}, a Feasibility-Aware and Robust Direct Preference Optimization algorithm. Through training-time preference alignment, \mbox{FAR-DPO} shifts a model's generative distribution toward cyclic peptides that combine structural feasibility with better binding affinity. Our main contributions are as follows:

\begin{itemize}
    \item \textbf{Feasibility-aware preference data construction strategy for cyclic peptides.} From the same balanced set of 12{,}000 pockets drawn from CPCore, a high-quality subset of the CPSea cyclic peptide--protein complex dataset \cite{yang2025cpsea}, the frozen CPSea-trained PepGLAD and PepFlow base models each generate 16 candidates per pocket, yielding 192{,}000 candidates per backbone. Using structural feasibility screening and within-pocket multi-objective comparison, we construct 90{,}408 and 92{,}099 preference pairs for PepGLAD and PepFlow, respectively.
    \item \textbf{Difficulty-aware robust preference optimization.} For each backbone, we further estimate pocket difficulty from its own frozen reference-model generations and assign fixed difficulty-group labels to the resulting preference pairs. We introduce a group-robust objective into direct preference optimization. The objective dynamically adjusts the optimization weights of difficulty groups according to their preference losses, placing greater emphasis on high-loss groups to help reduce the risk that aggregate objectives mask performance differences across targets.
    \item \textbf{Validity across generative backbones.} We evaluate \mbox{FAR-DPO} across two cyclic peptide design backbones, PepGLAD and PepFlow. Under a fixed generation budget, \mbox{FAR-DPO} increases feasible-candidate yield across both methods while improving binding quality, supporting its applicability across distinct generative architectures.
\end{itemize}

\section{Related Work}

\subsection{Cyclic Peptide Generation}

Physics-based sampling and design enabled stable constrained peptides before deep generative models \cite{bhardwaj2016hyperstable,hosseinzadeh2017macrocycles}. Recently, PepGLAD~\cite{kong2024pepglad}, PepFlow~\cite{li2024pepflow}, and PPFlow~\cite{lin2024ppflow} have generated target-conditioned peptide sequences and structures using geometric diffusion or flow matching for linear peptide generation. AfCycDesign~\cite{rettie2025afcyc} instead adapts positional encodings, whereas RFpeptides~\cite{rettie2025rfpeptides} and CPSDE~\cite{zhou2025cpsde} generate full-atom macrocycles. DiffPepBuilder~\cite{wang2024diffpepbuilder} adds disulfide bonds after generation, CP-Composer~\cite{jiang2025cpcomposer} composes geometric constraints, and APCyc~\cite{zhao2026apcyc} selects cyclization types and sites. CYC\_BUILDER~\cite{wang2025cycbuilder} uses search and reinforcement learning for fragment assembly, whereas GeoCycler~\cite{zhang2026geocycler} uses reward-based fine-tuning for closure constraints. \mbox{FAR-DPO} also leaves the generator unchanged, but uses preference-based post-training to increase the number of candidates that jointly satisfy structural and biophysical requirements.

\subsection{Preference Alignment for Generative Models}

Likelihood-based training does not guarantee specific design objectives, motivating reinforcement learning, multi-objective search, and guided generation \cite{olivecrona2017molecular,jin2020multiobjective,tang2024peptune}. These approaches often require explicit rewards, property predictors, or online sampling. Instead, Direct Preference Optimization (DPO) learns from relative comparisons without a reward model \cite{rafailov2023dpo}. Diffusion-DPO and Linear-DPO extend this principle to diffusion and flow-matching generators \cite{wallace2024diffusiondpo,li2026lineardpo}. Preference alignment has been applied to antibodies \cite{zhou2024abdpo,ren2025abnovo}, structure-based drug design \cite{cheng2025decompdpo,huang2025molform}, and peptides \cite{zhang2026pepald}. Most methods target a particular representation, model family, or property. \mbox{FAR-DPO} instead combines feasibility-gated, multi-objective preference pairs within each pocket and employs a difficulty-aware objective for optimization across targets.

\subsection{Group-Robust Optimization}

Average-loss training can mask persistent errors in difficult groups. DRO optimizes risk under an adversarial distribution \cite{duchi2021uniform}; Group DRO dynamically emphasizes high-loss predefined groups \cite{sagawa2020groupdro}, while TERM uses exponential tilting to balance average and tail performance \cite{li2021term}. These methods have been studied mainly in supervised learning. In contrast, \mbox{FAR-DPO} defines groups by pocket difficulty and applies group-robust weighting during preference-based post-training for target-conditioned generation.

\section{Method}

\subsection{Problem Formulation and Method Overview}

Given a protein binding pocket $P$, a pretrained pocket-conditioned reference generator $\pi_{\mathrm{ref}}(x\mid P)$ generates cyclic peptide candidates $x$, each comprising a peptide sequence and its full-atom structure. Without modifying the underlying generative architecture, \mbox{FAR-DPO} post-trains the model using offline preference pairs to obtain a policy $\pi_\theta(x\mid P)$ that favors preferred candidates over dispreferred ones relative to the frozen reference model.

\begin{figure*}[t]
\centering
\includegraphics[width=\textwidth]{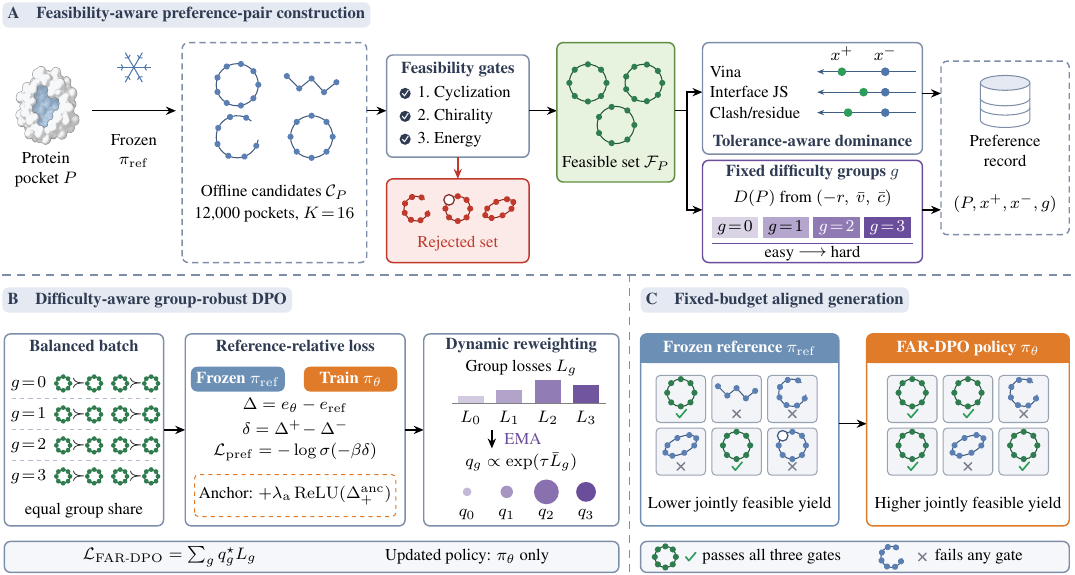}
\caption{Overview of \mbox{FAR-DPO}. (A) Preference pairs are constructed through feasibility screening and tolerance-aware dominance comparisons within each pocket and assigned predefined pocket-difficulty group labels. (B) Training combines group-balanced sampling, reference-relative preference optimization with preferred-candidate anchoring, and dynamic group reweighting based on group losses. (C) Under identical pockets, sampling budgets, and screening criteria, the \mbox{FAR-DPO} policy yields more jointly feasible candidates than the frozen reference.}
\label{fig:overview}
\end{figure*}

As illustrated in Fig.~\ref{fig:overview}, \mbox{FAR-DPO} first constructs within-pocket preference pairs under feasibility constraints and assigns them fixed difficulty-group labels. It then performs preference-based training by combining group-balanced sampling, preferred-candidate anchoring, and dynamic group reweighting. Finally, the trained policy generates candidates under a fixed sampling budget, and the jointly feasible candidate yield is evaluated. The resulting preference dataset is
$\mathcal D_{\mathrm{pref}}=\{(P_i,x_i^+,x_i^-,g_i)\}_{i=1}^{N}$, where $x_i^+$ and $x_i^-$ denote the preferred and dispreferred candidates, respectively, and $g_i$ is the difficulty group of their pocket.

\subsection{Feasibility-Aware Preference Construction}

\paragraph{Offline candidates and feasibility gating}
To construct offline data for preference-based post-training, we independently sample $K$ candidates for each protein binding pocket $P$ from the frozen reference model $\pi_{\mathrm{ref}}$ and denote the resulting set by $\mathcal C_P$. For each backbone, the candidate pool comprises 192{,}000 complexes generated from the same 12{,}000 CPCore pockets, with $K=16$ candidates per pocket. Generation uses the corresponding native peptide length, and candidate complexes undergo structural reconstruction and relaxation before feasibility screening and interface-quality evaluation.

Before defining preferences among candidates, we apply three eligibility gates covering cyclization geometry, residue chirality, and interaction energy. Candidates passing all three gates form the feasible set
\begin{equation}
\mathcal F_P =
\left\{
x\in\mathcal C_P:
G_{\mathrm{cyc}}(x)
G_{\mathrm{chir}}(x)
G_{\mathrm{ene}}(x)=1
\right\}.
\label{eq:feasible-set}
\end{equation}
Here, $G_{\mathrm{cyc}}$, $G_{\mathrm{chir}}$, and $G_{\mathrm{ene}}$ are binary indicators for cyclization geometry, residue chirality, and interaction energy, respectively. Following the CPSea evaluation protocol \cite{yang2025cpsea}, we use whether the $C_\beta$--$C_\beta$ distance between the N- and C-terminal residues lies in $[3,8]\,\text{\AA}$ as a proxy for cyclization geometry, while the chirality gate requires every residue to have the correct chirality. The energy gate requires Rosetta $\Delta G<0$.

\paragraph{Interface-quality evaluation}
For each $x\in\mathcal F_P$, we characterize interface quality using the Vina docking score $v(x)$, interface non-nativeness $j(x)$, and atomic clashes normalized by peptide length $c(x)$:
\begin{equation}
\mathbf m(x)=\big(v(x),j(x),c(x)\big).
\label{eq:metrics}
\end{equation}
The Vina score is computed using AutoDock Vina. Hydrophobic contacts, hydrogen bonds, and salt bridges at each candidate interface are identified using PLIP, and the interface non-nativeness score $j(x)$ is the Jensen--Shannon distance between the candidate's interface-interaction composition and a fixed natural-reference composition. All three metrics are minimized: lower $v(x)$, $j(x)$, and $c(x)$ indicate more favorable  binding, an interaction composition closer to the natural reference, and fewer atomic clashes per peptide residue, respectively. Preferences are defined only among candidates in $\mathcal F_P$ using $\mathbf m(x)$.

\paragraph{Scale-aware within-pocket dominance}
The three metrics have different units and numerical ranges, so comparisons based directly on raw differences would make preferences depend on metric scale. Let $\mathcal F=\bigcup_P\mathcal F_P$. We estimate a robust global scale for the $k$-th metric from all feasible candidates:
\begin{equation}
\sigma_k =
1.4826\,
\underset{x\in\mathcal F}{\operatorname{median}}
\left|
m_k(x)-
\underset{x'\in\mathcal F}{\operatorname{median}}m_k(x')
\right|.
\label{eq:global-mad}
\end{equation}
Here, $m_k(x)$ is the $k$-th component of $\mathbf m(x)$. The factor $1.4826$ makes the median absolute deviation (MAD) a robust scale estimate consistent with the standard deviation under a Gaussian distribution. Following the Pareto dominance principle in multi-objective optimization, we use $\sigma_k$ to define a tolerance-aware dominance relation within pocket $P$:
\begin{equation}
\begin{aligned}
x^+\succ_P x^-
\Longleftrightarrow\;&
\left[\forall k:\;
m_k(x^+)-m_k(x^-)\leq\kappa\sigma_k\right]\\
&{}\land
\left[\exists k:\;
m_k(x^+)-m_k(x^-)<-\kappa\sigma_k\right].
\end{aligned}
\label{eq:dominance}
\end{equation}
We set \(\kappa=0.5\).
Thus, $x^+$ cannot be worse than $x^-$ by more than the tolerance on any metric and must outperform it by more than the tolerance on at least one metric. All comparisons are restricted to the same pocket, preventing differences in pocket-level difficulty from being mistaken for differences in candidate quality. We rank pairs satisfying \eqref{eq:dominance} by their standardized advantage beyond the tolerance and retain up to eight of the highest-ranked preference pairs per pocket.

\subsection{Difficulty-Aware Pocket Grouping}

\paragraph{Pocket-level difficulty features}
Pockets vary in design difficulty for the reference model. To construct fixed group labels for group-robust training, we estimate the difficulty of each pocket from offline generations of the frozen reference model. For every pocket $P$ that admits at least one preference pair, we compute pocket-level features describing structural feasibility, binding affinity, and steric clashes. Specifically, $r(P)$ is the fraction of candidates in $\mathcal C_P$ that satisfy both the cyclization-geometry and residue-chirality gates, while $\bar v(P)$ and $\bar c(P)$ are the mean Vina score and normalized clash count, respectively, over $\mathcal F_P$.

\paragraph{Robust standardization and difficulty aggregation}
The three features have different numerical scales. For $f\in\{-r,\bar v,\bar c\}$, we apply MAD-based robust standardization across all pockets that admit preference pairs:
\begin{equation}
z_f(P)=
\operatorname{clip}
\left(
\frac{
f(P)-\operatorname{median}_{P'}f(P')
}{
1.4826\,
\operatorname{MAD}_{P'}\!\left(f(P')\right)
},
-4,4
\right).
\label{eq:difficulty-standardize}
\end{equation}
Here, $P'$ ranges over all pockets that admit preference pairs. Clipping to $[-4,4]$ limits the influence of extreme pockets on the difficulty score. We then aggregate the standardized features into a pocket-difficulty score:
\begin{equation}
D(P)=
0.2\,z_{-r}(P)+
0.6\,z_{\bar v}(P)+
0.2\,z_{\bar c}(P).
\label{eq:difficulty}
\end{equation}

\paragraph{Difficulty-group assignment}
A larger $D(P)$ indicates greater pocket design difficulty. We partition pockets by quartiles of $D(P)$ into four predefined difficulty groups, $g(P)\in\{0,1,2,3\}$, where groups 0 and 3 correspond to the easiest and hardest quartiles, respectively. All preference pairs from the same pocket share the same $g(P)$. Difficulty groups are fixed in advance from the offline candidate pool and remain unchanged throughout training. In the evaluation, Q1--Q4 denote analogous easiest-to-hardest quartiles constructed and frozen separately for each generator.

After assigning difficulty groups, we split each backbone-specific dataset into training and validation sets by pocket, ensuring that no pocket appears in both splits. The resulting datasets contain 90{,}408 preference pairs for PepGLAD and 92{,}099 preference pairs for PepFlow.

\subsection{Group-Robust Direct Preference Optimization}

\paragraph{Error surrogate and reference-relative change}
The diffusion and flow-matching generators considered here do not provide tractable conditional log-likelihoods in the form required by standard DPO. Following Diffusion-DPO, \mbox{FAR-DPO} therefore constructs a preference-optimization surrogate from the native prediction error of each generative backbone \cite{wallace2024diffusiondpo}. Let $e_\theta(x;\xi)$ and $e_{\mathrm{ref}}(x;\xi)$ denote the prediction errors of the policy and frozen reference model, respectively, for candidate $x$ under stochastic training variables $\xi$, which include the timestep and random noise. For PepGLAD, the error combines sequence-latent and structure-latent prediction losses using the base model's weighting. For PepFlow, it is the equally weighted sum of translation and rotation flow-matching errors. We define the policy's reference-relative error change as
\begin{equation}
\Delta_\theta(x;\xi)
=e_\theta(x;\xi)-e_{\mathrm{ref}}(x;\xi).
\label{eq:error-change}
\end{equation}
For a given candidate, the policy and reference model share the same $\xi$. The coupling of stochastic variables between the two candidates is backbone-specific. A negative $\Delta_\theta(x;\xi)$ indicates that the policy reduces the candidate's prediction error relative to the reference model.

\paragraph{Pairwise preference and anchoring}
For preference pair $i=(x_i^+,x_i^-)$, let $\xi_i^+$ and $\xi_i^-$ denote the stochastic training variables associated with its two candidates. The pairwise preference loss is
\begin{equation}
\begin{aligned}
\delta_i
&=\Delta_\theta(x_i^+;\xi_i^+)
-\Delta_\theta(x_i^-;\xi_i^-),\\
\mathcal L_{\mathrm{pref}}^{(i)}
&=-\log \operatorname{sigmoid}(-\beta\delta_i).
\end{aligned}
\label{eq:preference-loss}
\end{equation}
where $\beta$ controls the strength of the pairwise preference signal. To prevent the relative margin from improving by allowing $x_i^+$ itself to deteriorate relative to the reference model, let $\Delta_\theta^{\mathrm{anc}}$ denote the reference-relative change in the backbone-specific anchoring error. We augment the preference loss with
\begin{equation}
\ell_i =
\mathcal L_{\mathrm{pref}}^{(i)}
+\lambda_{\mathrm a}
\operatorname{ReLU}
\left(
\Delta_\theta^{\mathrm{anc}}(x_i^+;\xi_i^+)
\right),
\label{eq:anchor}
\end{equation}
where $\lambda_{\mathrm a}$ controls the anchoring strength.

\paragraph{Difficulty-group losses and dynamic weighting}
Motivated by the emphasis on high-loss groups in Group DRO \cite{sagawa2020groupdro}, each mini-batch is constructed by balanced sampling across the four difficulty groups. Let $\mathcal B_g$ denote the preference pairs from difficulty group $g$ in the current batch. Their mean group loss is
\begin{equation}
L_g(\theta)=
\frac{1}{|\mathcal B_g|}
\sum_{i\in\mathcal B_g}\ell_i.
\label{eq:group-loss}
\end{equation}
To reduce mini-batch variance, we maintain an exponential moving average of each group loss with decay coefficient $\rho$:
\begin{equation}
\bar L_g
\leftarrow
\rho\,\bar L_g+(1-\rho)L_g.
\label{eq:group-ema}
\end{equation}
We then solve for a KL-regularized adversarial group distribution based on $\bar L_g$ \cite{duchi2021uniform}:
\begin{equation}
\begin{aligned}
\mathbf q^\star
=\underset{\mathbf q\in\Delta_4}{\arg\max}\;
&\left[
\sum_{g=0}^{3}q_g\bar L_g
-\frac{1}{\tau}
\operatorname{KL}(\mathbf q\,\|\,\mathbf u)
\right],\\
q_g^\star
=\;&
\frac{\exp(\tau\bar L_g)}
{\sum_{h=0}^{3}\exp(\tau\bar L_h)}.
\end{aligned}
\label{eq:group-weights}
\end{equation}
Here, $\Delta_4$ is the probability simplex over the four difficulty groups, $\mathbf u=(1/4,\ldots,1/4)$ is the uniform prior, and $\tau$ controls how strongly the group weights concentrate. The final training objective aggregates the current group losses using these dynamic weights:
\begin{equation}
\mathcal L_{\mathrm{FAR\text{-}DPO}}(\theta)
=\sum_{g=0}^{3}q_g^\star L_g(\theta).
\label{eq:far-dpo}
\end{equation}

\section{Experiments}

Taking PepGLAD and PepFlow as the pretrained backbones, we assess whether \mbox{FAR-DPO} increases the yields of feasible and jointly qualified candidates for cyclic peptide design. We also evaluate its performance on challenging targets and under limited sampling budgets. Further analysis examines how different optimization choices contribute to the results.

\begin{table*}[t]
\caption{Overall Performance on the 56-Target LNR Benchmark}
\label{tab:overall}
\centering
\setlength{\tabcolsep}{3pt}
\begin{tabular}{cccccccc}
\toprule
\multirow{2}{*}{Group} & \multirow{2}{*}{Metric} & \multicolumn{1}{c}{\multirow{2}{*}{CP-Composer}} & \multicolumn{1}{c}{\multirow{2}{*}{DiffPepBuilder}} & \multicolumn{1}{c}{\multirow{2}{*}{PepFlow}} & \multicolumn{1}{c}{\mbox{FAR-DPO}} & \multicolumn{1}{c}{\multirow{2}{*}{PepGLAD}} & \multicolumn{1}{c}{\mbox{FAR-DPO}} \\
 &  &  &  &  & \multicolumn{1}{c}{(PepFlow)} &  & \multicolumn{1}{c}{(PepGLAD)} \\
\midrule
\multirow{4}{*}{Success}
& Cyclization (\%) $\uparrow$ & 70.3 & 92.5 & 93.7 & \textbf{94.0} & 81.5 & 90.2 \\
& Chirality (\%) $\uparrow$ & 64.3 & 74.6 & 78.4 & 79.6 & 81.4 & \textbf{85.5} \\
& Energy pass (\%) $\uparrow$ & 49.7 & \textbf{75.1} & 63.2 & 64.8 & 70.4 & 74.4 \\
& Final Success (\%) $\uparrow$ & 24.2 & 52.2 & 48.0 & 49.6 & 46.9 & \textbf{57.8} \\
\midrule
\multirow{2}{*}{Affinity}
& Rosetta dG best/target \(\downarrow\) & $-$18.5 & \textbf{\boldmath$-$23.3} & $-$20.7 & $-$21.1 & $-$23.0 & \textbf{\boldmath$-$23.3} \\
& Vina best/target \(\downarrow\) & $-$5.10 & $-$5.90 & $-$6.77 & \textbf{\boldmath$-$6.82} & $-$6.30 & $-$6.74 \\
\midrule
\multirow{2}{*}{Rama.}
& Allowed (\%) $\uparrow$ & 73.8 & \textbf{86.9} & 83.4 & 83.1 & 72.6 & 71.5 \\
& Favored (\%) $\uparrow$ & 41.6 & \textbf{65.4} & 58.9 & 58.4 & 40.7 & 40.1 \\
\midrule
\multirow{2}{*}{scRMSD}
& Mainchain (\AA) $\downarrow$ & 2.78 & \textbf{2.31} & 2.37 & 2.38 & 2.68 & 2.75 \\
& Disulfide (\AA) $\downarrow$ & 2.60 & \textbf{2.34} & 2.74 & 2.73 & 2.63 & 2.59 \\
\midrule
\multirow{2}{*}{Physicochem.}
& GRAVY $<0$ (\%) $\uparrow$ & 57.8 & \textbf{83.9} & 60.8 & 61.3 & 78.7 & 75.3 \\
& logP $>-6$ (\%) $\uparrow$ & 55.3 & \textbf{92.3} & 46.8 & 46.0 & 67.5 & 68.0 \\
\midrule
\multirow{2}{*}{Interface}
& Interface JS $\downarrow$ & 0.184 & 0.166 & 0.181 & 0.178 & 0.160 & \textbf{0.143} \\
& Clash/residue $\downarrow$ & 0.394 & 0.342 & 0.277 & 0.264 & 0.342 & \textbf{0.259} \\
\midrule
\multirow{2}{*}{Distribution}
& Diversity $\uparrow$ & \textbf{0.759} & 0.337 & 0.573 & 0.565 & 0.596 & 0.607 \\
& Novelty $\uparrow$ & 0.122 & 0.120 & 0.137 & \textbf{0.140} & 0.121 & 0.127 \\
\bottomrule
\end{tabular}

\end{table*}

\subsection{Experimental Setup}
\label{sec:experimental-setup}

\paragraph{Data and baselines}
We evaluate all methods on the same 56-target subset of the large non-redundant (LNR) dataset \cite{tsaban2022harnessing} curated in CPSea \cite{yang2025cpsea}, using 100 raw candidates per target and the CPSea post-processing and evaluation protocol. We compare PepGLAD \cite{kong2024pepglad} and PepFlow \cite{li2024pepflow} with their \mbox{FAR-DPO} counterparts, with DiffPepBuilder \cite{wang2024diffpepbuilder} and CP-Composer \cite{jiang2025cpcomposer} as external baselines. CP-Composer uses head-to-tail cyclization with a classifier-free guidance weight of 5.0.

\paragraph{Evaluation metrics}
In addition to the CPSea evaluation metrics, we assess interface quality using Interface JS and Clash/residue, which measure the Jensen--Shannon distance between a candidate's interface-interaction composition and a fixed natural-reference composition, and the number of inter-chain atomic clashes normalized by the number of peptide residues, respectively. Candidates passing the cyclization, chirality, and energy gates defined above are referred to as Final candidates. For each backbone and target, we compute a reference median for each of the three metrics---Vina, Interface JS, and Clash/residue---from the corresponding base model's Final candidates with valid values for all three metrics. Both jointly qualified (JQ) and MAD-qualified candidates must belong to the Final pool. JQ candidates must have Vina and Interface JS values no higher than their respective reference medians and exhibit no inter-chain heavy-atom clashes. MAD-qualified candidates must improve at least one of the three metrics beyond its prescribed tolerance relative to the corresponding reference median, without degrading any of the other metrics beyond its respective tolerance. Both backbones use the same fixed evaluation tolerances of \((0.575871,\,0.045096,\,0.088250)\) for these three metrics, respectively. We report the number of JQ- and MAD-qualified candidates per target and their proportions within the respective evaluable Final candidate pools. Low-budget performance is evaluated using Pass@\(K\) and Raw@10. Pass@\(K\) is the probability of obtaining at least one Final candidate when \(K\) candidates are sampled uniformly without replacement from the 100 raw candidates for each target. Raw@10 is estimated separately for Vina and Rosetta dG using 10,000 candidate-sampling trials per target, each drawing 10 raw candidates. We record the best score among Final candidates with valid scores in each trial and average these scores only over trials containing at least one such candidate.

\paragraph{Difficulty groups and statistical analysis}
We apply the difficulty features and scoring procedure in \eqref{eq:difficulty-standardize}--\eqref{eq:difficulty} separately to each base model's test generations and partition the targets into Q1--Q4, with 14 targets per group. Missing continuous features are imputed using the corresponding median across the remaining test targets for the same backbone. Q4 is the hardest group, and its Final Success is denoted as Q4 Final. Each base model and its \mbox{FAR-DPO} counterpart share fixed groups constructed without fine-tuned outputs; these test groups are distinct from the training pocket groups. Affinity and interface metrics, Q4 Final, Pass@$K$, Raw@10, and hits/target are target-macro averages. Stage pass rates, structural and physicochemical metrics, and JQ/MAD rates are pooled over their respective evaluable candidate sets; diversity and novelty are global. We compute 95\% percentile confidence intervals from 10,000 target-level bootstrap resamples \cite{efron1979bootstrap}, paired between each base model and its \mbox{FAR-DPO} counterpart. Within-group analyses resample the corresponding 14 targets, and pooled rates are recomputed after each resampling.

\subsection{Overall Performance and Joint Quality}

To determine whether \mbox{FAR-DPO} improves feasible-candidate yield across generative backbones and to examine its effects on other quality dimensions, we compare the base models with their \mbox{FAR-DPO} counterparts, as shown in Table~\ref{tab:overall}. \mbox{FAR-DPO}--PepGLAD increases Final Success from 46.9\% to 57.8\%, while \mbox{FAR-DPO}--PepFlow increases it from 48.0\% to 49.6\%. Cyclization, chirality, and energy pass rates improve for both backbones. Vina best/target improves from $-6.30$ and $-6.77$ to $-6.74$ and $-6.82$, respectively, and Rosetta dG also improves further. In addition, \mbox{FAR-DPO} reduces Interface JS and Clash/residue for both backbones, while changes in conformational, physicochemical, and generative-distribution metrics are generally modest. These results demonstrate that \mbox{FAR-DPO} improves feasible-candidate yield and binding quality across different generative backbones under a fixed generation budget, with limited trade-offs in several secondary quality dimensions.

To further examine whether increased feasible-candidate yield is accompanied by improved joint quality, we report the numbers of JQ- and MAD-qualified candidates and their proportions within the respective evaluable Final candidate pools (Table~\ref{tab:joint}). For PepGLAD, JQ hits/target increase from 8.70 to 16.54, with the corresponding rate rising from 18.61\% to 28.73\%; MAD hits/target increase from 13.95 to 24.54, with the corresponding rate rising from 29.84\% to 42.63\%. All four metrics also improve for PepFlow. These results show that \mbox{FAR-DPO} increases both the number of jointly qualified candidates and their proportion within the respective evaluable Final candidate pools.

\begin{table}[t]
\caption{Jointly Qualified Yield and MAD Robustness}
\label{tab:joint}
\centering
\setlength{\tabcolsep}{2.5pt}
\begin{tabular}{ccccc}
\toprule
Metric & \multicolumn{1}{c}{PepFlow} & \multicolumn{1}{c}{\shortstack{\mbox{FAR-DPO}\\(PepFlow)}} & \multicolumn{1}{c}{PepGLAD} & \multicolumn{1}{c}{\shortstack{\mbox{FAR-DPO}\\(PepGLAD)}} \\
\midrule
JQ hits/target $\uparrow$ & 10.29 & \textbf{11.00} & 8.70 & \textbf{16.54} \\
JQ rate in Final (\%) $\uparrow$ & 21.46 & \textbf{22.22} & 18.61 & \textbf{28.73} \\
MAD hits/target $\uparrow$ & 14.27 & \textbf{15.43} & 13.95 & \textbf{24.54} \\
MAD rate in Final (\%) $\uparrow$ & 29.77 & \textbf{31.17} & 29.84 & \textbf{42.63} \\
\bottomrule
\end{tabular}

\end{table}

\subsection{Target Difficulty and Low-Budget Sampling}

\begin{figure}[!t]
\centering
\includegraphics[width=\columnwidth]{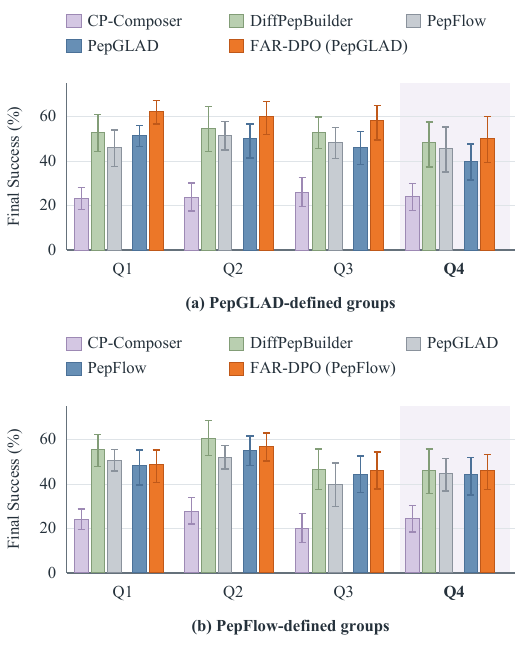}
\caption{Final Success across fixed difficulty groups. (a) PepGLAD-defined groups; (b) PepFlow-defined groups. Difficulty increases from Q1 to Q4, with 14 targets per group. Error bars indicate 95\% bootstrap confidence intervals.}
\label{fig:difficulty}
\end{figure}

To determine whether the overall gains extend across target difficulty, particularly to the hardest targets, we compare Final Success across the four frozen difficulty groups defined separately for the two generative backbones (Fig.~\ref{fig:difficulty}). \mbox{FAR-DPO} improves Final Success in all four difficulty groups for both generators. In Q4, \mbox{FAR-DPO}--PepGLAD improves Final Success from 40.00\% to 50.36\%, while \mbox{FAR-DPO}--PepFlow improves it from 44.14\% to 46.29\%. These results show that the gains span the full difficulty range and extend to the most challenging targets for both models.

\begin{figure}[!t]
\centering
\includegraphics[width=\columnwidth]{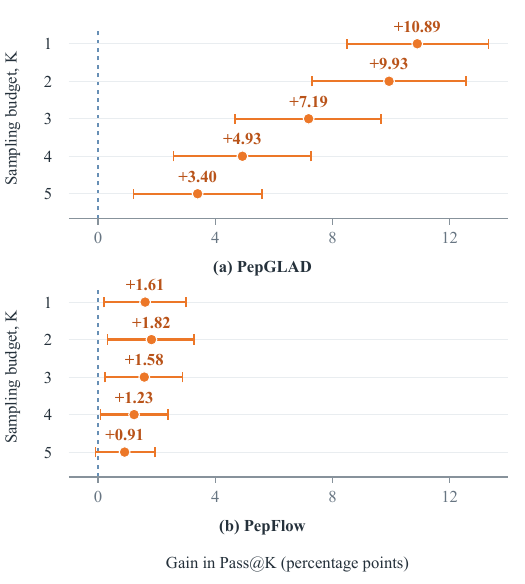}
\caption{Pass@$K$ gains from \mbox{FAR-DPO} under low sampling budgets. (a) PepGLAD; (b) PepFlow. Points and error bars indicate mean gains over the corresponding base model in percentage points and their 95\% paired bootstrap confidence intervals.}
\label{fig:pass-k}
\end{figure}

Because practical design campaigns often operate under limited generation budgets, we further examine the probability of obtaining at least one Final candidate at different sampling budgets (Fig.~\ref{fig:pass-k}). \mbox{FAR-DPO} improves Pass@$K$ for both generative backbones. \mbox{FAR-DPO}--PepGLAD increases Pass@1 from 46.89\% to 57.79\% and Pass@5 from 91.52\% to 94.92\%; \mbox{FAR-DPO}--PepFlow increases Pass@1 from 47.96\% to 49.57\% and Pass@5 from 91.63\% to 92.54\%. The advantage is most pronounced in the low-budget regime, indicating that \mbox{FAR-DPO} produces feasible designs with fewer generations and thereby improves practical sampling efficiency.

\subsection{Analysis of Optimization Variants}

\begin{table*}[t]
\caption{Optimization Variants on PepGLAD}
\label{tab:variants}
\centering
\setlength{\tabcolsep}{1.95pt}
\begin{tabular}{@{}cccccccccc@{}}
\toprule
& \multicolumn{3}{c}{Components} & \multicolumn{2}{c}{Feasibility} & \multicolumn{4}{c}{Quality} \\
\cmidrule(lr){2-4}\cmidrule(lr){5-6}\cmidrule(l){7-10}
Variant &
\shortstack{Group-balanced\\sampling} &
\shortstack{Dynamic group\\reweighting} &
\shortstack{Anchoring\\$\lambda_{\mathrm a}=0.2$} &
\shortstack{Final Success\\(\%) $\uparrow$} &
\shortstack{Q4 Final\\(\%) $\uparrow$} &
\shortstack{Raw@10\\Vina $\downarrow$} &
\shortstack{Raw@10\\Rosetta dG $\downarrow$} &
\shortstack{Interface\\JS $\downarrow$} &
\shortstack{Clash/residue\\$\downarrow$} \\
\midrule
Base PepGLAD & -- & -- & -- & 46.89 & 40.00 & $-$5.123 & $-$16.830 & 0.1596 & 0.3420 \\
DPO & $\times$ & $\times$ & $\times$ & 46.14 & 37.43 & $-$5.604 & $-$16.755 & \textbf{0.1424} & 0.2805 \\
Balanced DPO & $\checkmark$ & $\times$ & $\times$ & 49.21 & 43.86 & $-$5.669 & $-$17.554 & 0.1437 & 0.2844 \\
Balanced + anchoring & $\checkmark$ & $\times$ & $\checkmark$ & 50.32 & 43.14 & $-$5.578 & \textbf{\boldmath$-$17.638} & 0.1429 & 0.2832 \\
Group-DRO DPO & $\checkmark$ & $\checkmark$ & $\times$ & 52.77 & 46.71 & \textbf{\boldmath$-$5.726} & $-$16.954 & 0.1429 & \textbf{0.2549} \\
\textbf{\mbox{FAR-DPO}} & $\checkmark$ & $\checkmark$ & $\checkmark$ & \textbf{57.79} & \textbf{50.36} & $-$5.579 & $-$17.383 & 0.1425 & 0.2592 \\
\bottomrule
\end{tabular}
\end{table*}

We compare optimization variants on PepGLAD to examine the contributions of group-balanced sampling, dynamic group reweighting, and preferred-candidate anchoring to feasible-candidate yield (Table~\ref{tab:variants}). Full \mbox{FAR-DPO} achieves the highest overall and Q4 Final Success among the evaluated configurations, reaching 57.79\% and 50.36\%, respectively. Compared with standard DPO, group-balanced sampling improves both success rates, and incorporating dynamic group reweighting yields further gains. Adding preferred-candidate anchoring to this configuration further increases overall and Q4 Final Success by 5.02 and 3.65 percentage points, respectively. In addition, full \mbox{FAR-DPO} outperforms the base model on all four quality metrics: Raw@10 Vina, Raw@10 Rosetta dG, Interface JS, and Clash/residue. These results support combining the three components to increase feasible-candidate yield overall and on challenging targets while improving candidate quality relative to the base model.

\subsection{Three-Dimensional Case Studies}

\begin{figure}[!t]
\centering
\includegraphics[width=\columnwidth]{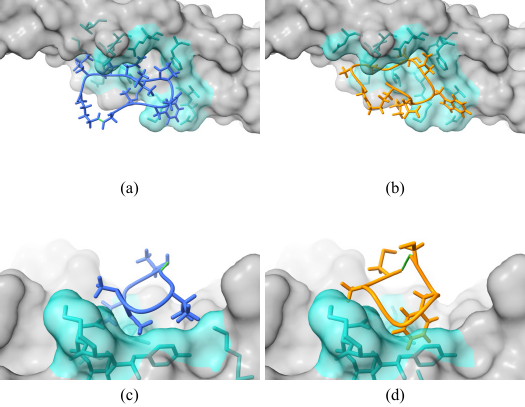}
\caption{Representative complexes for 5LM1: (a) PepGLAD and (b) \mbox{FAR-DPO}--PepGLAD; and 3BRH: (c) PepFlow and (d) \mbox{FAR-DPO}--PepFlow. Base-model and \mbox{FAR-DPO} peptides are blue and orange; receptors, interface residues within 4 \AA{}, and cyclization bonds are gray, cyan, and green, respectively. Each model independently generates candidates; the representative has the lowest Vina score among Final-pass candidates within the corresponding cyclization topology.}
\label{fig:structures}
\end{figure}

Fig.~\ref{fig:structures} compares independently generated Final-pass designs from each model and its \mbox{FAR-DPO} counterpart. Under 100 generations per model, \mbox{FAR-DPO}--PepGLAD increases the number of Final candidates for 5LM1 from 34 to 55; the representative designs have Vina scores of $-4.59$ and $-6.61$ and interface-contact counts of 9 and 11, respectively. For 3BRH, \mbox{FAR-DPO}--PepFlow increases the number of Final candidates from 47 to 53, while the representative Vina score and Rosetta dG improve from $-4.35$ to $-4.58$ and from $-9.72$ to $-13.13$. These cases illustrate improvements in feasible-candidate yield and representative predicted binding quality across both backbones.

\section{Conclusion}

We introduced \mbox{FAR-DPO}, an architecture-agnostic method for target-specific cyclic peptide generation under multiple structural and biophysical constraints. \mbox{FAR-DPO} combines feasibility-gated, tolerance-aware within-pocket preferences with difficulty-aware group-robust optimization and preferred-candidate anchoring. Experiments with PepGLAD and PepFlow show higher jointly feasible candidate yield, with gains extending to challenging targets and low-budget sampling while retaining competitive binding quality. The optimization-variant analysis supports combining group-balanced sampling, dynamic group reweighting, and anchoring. These results suggest that preference-based post-training can improve joint feasibility and robustness across targets without modifying the generative architecture.

\section*{Acknowledgment}

This study was supported by the National Key R\&D Program of China (2025YFF0515300, 2025YFF0515301, and 2024YFF1207103), and the Shenzhen Medical Research Fund (E250200620, E250200622 to G.W.), the National Natural Science Foundation of China (T2522008 and 62501406 to G.W.). Additional support was provided by the Scientific Research Innovation Capability Support Project for Young Faculty (SRICSPYF-ZY2025015 to G.W.), and the Fundamental Research Funds for Beijing University of Posts and Telecommunications (Grant 2025AI4S18 to G.W.). This work was also supported by the New Cornerstone Science Foundation through the XPLORER PRIZE, and funded by the Beijing University of Posts and Telecommunications-China Mobile Communications Group Co., Ltd. Joint Institute. The funders had no role in study design, data collection and analysis, decision to publish, or preparation of the manuscript.

\bibliographystyle{IEEEtran}
\bibliography{far_dpo_refs_ieee}

\end{document}